\documentclass[lettersize,journal]{IEEEtran}
\usepackage{amsmath,amsfonts}
\usepackage{algorithmic}
\usepackage{algorithm}
\usepackage{array}
\usepackage[caption=false,font=normalsize,labelfont=sf,textfont=sf]{subfig}
\usepackage[labelfont=bf]{caption}
\usepackage{textcomp}
\usepackage{stfloats}
\usepackage{url}
\usepackage{verbatim}
\usepackage{graphicx}
\usepackage{cite}
\usepackage{authblk}
\usepackage{booktabs}
\usepackage{hyperref} 
\usepackage{cleveref} 
\usepackage[switch]{lineno}
\usepackage{float}
\usepackage[dvipsnames]{xcolor}
\definecolor{myblue}{rgb}{0.239,0.553,0.565}

\Crefformat{figure}{#2Fig.~#1#3}
\Crefformat{equation}{#2Eq.~#1#3}

\newcommand{\norm}[1]{\left\lVert#1\right\rVert}

\usepackage{color,soul}

\begin{document}

\title{Spline-FRIDA: Towards Diverse, Humanlike Robot Painting Styles with a Sample-Efficient, Differentiable Brush Stroke Model}

\author[*]{Lawrence Chen}
\author[*]{Peter Schaldenbrand}
\author[*]{Tanmay Shankar}
\author[*]{Lia Coleman}
\author[*]{Jean Oh}
\affil[*]{The Robotics Institute, Carnegie Mellon University}


\maketitle

\footnotetext[1]{This work has been submitted to the IEEE for possible publication. Copyright may be transferred without notice, after which this version may no longer be accessible.}

\section{Abstract}

A painting is more than just a picture on a wall; a painting is a \textit{process} comprised of many intentional brush strokes, the shapes of which are an important component of a painting's overall style and message. Prior work in modeling brush stroke trajectories either does not work with real-world robotics or is not flexible enough to capture the complexity of human-made brush strokes. 
In this work, we introduce Spline-FRIDA which can model complex human brush stroke trajectories. 
This is achieved by recording artists drawing using motion capture, modeling the extracted trajectories with an autoencoder, and introducing a novel brush stroke dynamics model to the existing robotic painting platform FRIDA. 
We conducted a survey and found that our open-source Spline-FRIDA approach successfully captures the stroke styles in human drawings and that Spline-FRIDA's brush strokes are more human-like, improve semantic planning, and are more artistic compared to existing robot painting systems with restrictive B\'ezier curve strokes.
\vspace{-1em}
\section{Introduction}

\begin{figure}[!h]
    \centering
    \includegraphics[width=\linewidth]{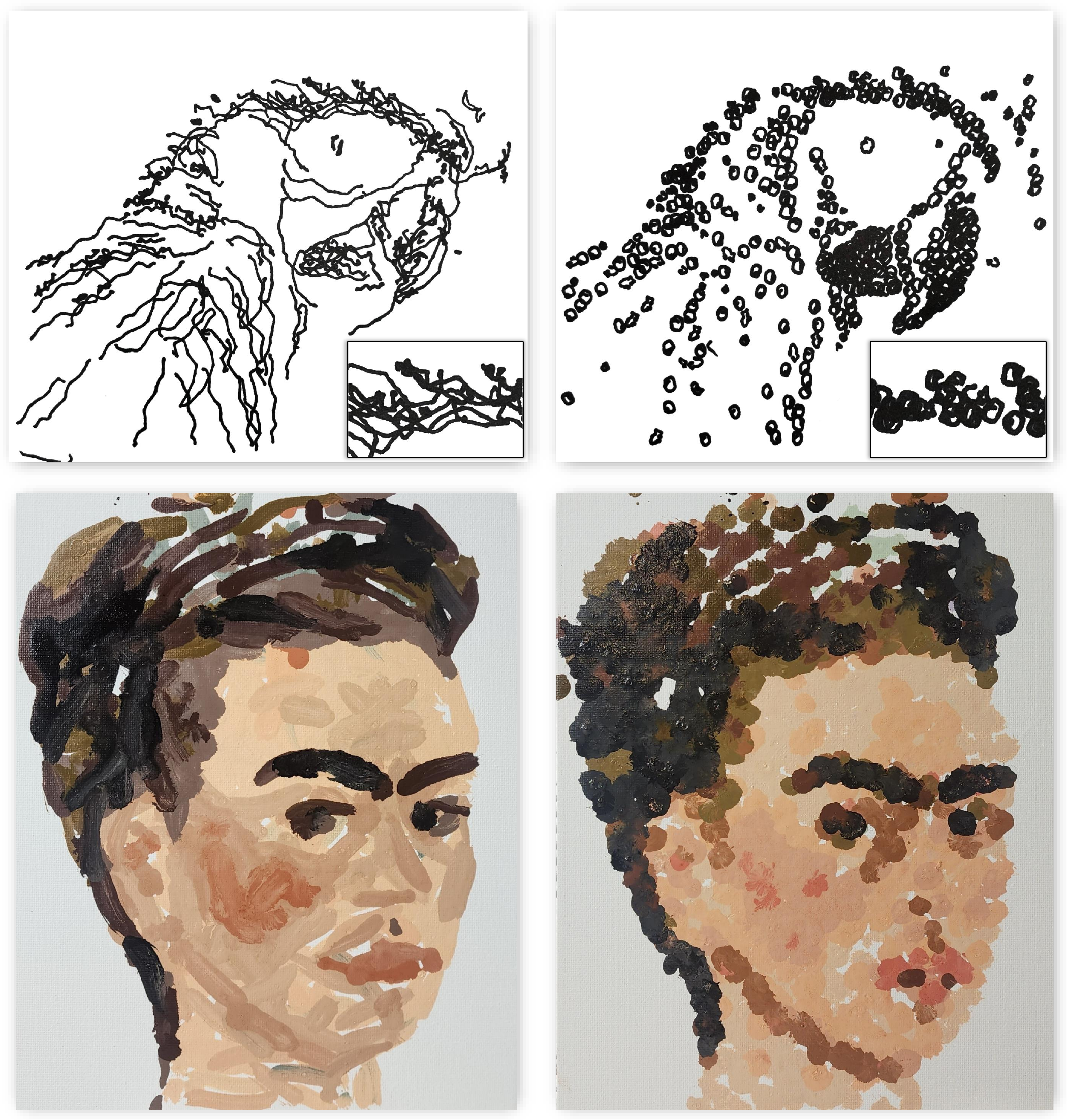}
    \caption{\small \textbf{Spline-FRIDA drawings and paintings in different styles.} These are two pairs of artworks of made by our system. The left paintings use longer, zig-zagging strokes, while the right ones are composed of small circles and dots. While each pair depicts the same content, the stroke style vastly changes the appearance and vibe of each work.}
    \label{fig:style_in_real_paintings}
    \vspace{-1.5em}
\end{figure}

Paintings and drawings are used to convey messages of emotion, cultural values, and shared experiences. While these aspects can be conveyed by the objects or subjects within the painting, style is perhaps just as important to expressing those messages \cite{robertson1967-formAndContent, schaldenbrand2022styleclipdraw}. 
In particular, patterns in the shapes of individual strokes within a painting can contribute to the overall style and aesthetic of an artwork.
Some examples can be seen in \Cref{fig:style_in_real_paintings}, with the left drawings using long strokes and the right drawings using small, circular ones. In both cases, the stroke shapes are crucial for defining the painting's style and therefore the expression of the message that the artist intends to convey. 

Furthermore, if robots are to support humans in the creation of artwork, it is important for the robot to have flexible styles of strokes for the user to specify either through choice or demonstration. Many artists do not wish to automate the artistic process~\cite{gebru2023artAiImpact, jansen2021coDrawWorkflows}, but some are open to co-creative assistants~\cite{bateman2021creating, davis2016DrawingApprentice, lawton2023reframer, lee2024adversarialRobot}. The work operates on the assumption that giving more creative control to a user co-creating with a robot over the style of the image, allows them to feel more ownership over the artwork that they create with the robot.

Prior work has mostly focused on planning paintings using basic stroke representations such as B\'ezier curves~\cite{learningtopaint,frida}\ and only fixating on the style of the overall image~\cite{schaldenbrand2022styleclipdraw, stylizedneuralpainting}. 
In this paper, we focus on how intra-stroke style control can be implemented.


Our work uses motion capture to record human drawings with real-world brushes and markers on paper. We model these recorded trajectories with an autoencoder, TrajVAE.
We also introduce a novel brush stroke dynamics model, Traj2Stroke, which predicts the 2d outline of stroke given its trajectory.

To summarize, our main contributions are as follows:
\begin{itemize}
    \item We present a successful approach to modeling human styles of brushstroke trajectories using variational autoencoders.
    \item We introduce Traj2Stroke, a method to differentiably render polylines with variable thickness. Compared to FRIDA's dynamics model, Traj2Stoke has a significantly smaller Sim2Real gap between simulated and real sharpie strokes.
    \item Our study gives evidence to suggest that the more flexible brush stroke trajectories can improve robot semantic planning and lead to more human-like paintings.
\end{itemize}

\section{Related Work}
Stroke-Based Rendering (SBR) involves arranging primitive shapes to create an image, often with the goal of replicating some target image. Some recent works use forward prediction methods, in which a neural network learns to output the next stroke to add \cite{contentmaskedloss, learningtopaint, painttransformer}, while others use optimization-based methods, where stroke parameters are passed through a differentiable rendering pipeline and optimized via backpropagation \cite{frida, stylizedneuralpainting, layerwise}.

\subsection{Stroke Primitives}
Most SBR research is focused on global planning and propose new algorithms to arrange stroke primitives. On the other hand, there has been little research into how the stroke primitives themselves should be defined. Some works use definitions that would be difficult to replicate on a physical robot. For instance, Learning to Paint defines strokes as translucent B\'ezier curves with arbitrary thicknesses~\cite{learningtopaint}. Schaldenbrand et al. found that when their system was restricted to outputting more realistic brush strokes by making them opaque and limiting the sizes, the quality of generated images suffered~\cite{contentmaskedloss}. Paint Transformer uses a mask of a brush stroke that can be transformed, resized, and recolored \cite{painttransformer}. This arbitrary sizing of strokes without loss of precision would be very difficult to implement in hardware.

Based on human art, many drawing tools, such as markers or brushes, can inherently be versatile and adaptable enough to produce a wide range of stroke styles. Specifically, altering the paths of individual strokes can result in diverse styles. This has been observed and researched extensively in the context of human handwriting replication \cite{helpinghand, handwriting}, but only to a lesser extent for drawings. We hope to further explore how to define stroke primitives by explicitly modeling the style of stroke trajectories used in a drawing.

\subsection{Differentiable Rendering}
In SBR, differentiable renderers are modules that take in stroke parameters and output a rendered image. They differ from traditional renderers in that gradients of the image with respect to the parameters can be obtained. Having access to such a module is a crucial assumption of many modern SBR planners.

Learning to Paint~\cite{learningtopaint} takes a reinforcement learning (RL) approach to SBR. Despite the fact that RL does not inherently require a differentiable environment, they found that using a differentiable renderer greatly boosted the system's performance and convergence rate compared to a model-free method. This is mainly because differentiable environment allows for end-to-end training of the RL agent. Paint Transformer~\cite{painttransformer} also makes use of differentiable rendering so that a loss can be backpropagated from the output image all the way back to its stroke predictor. These examples show that differentiable rendering can be useful even in methods that are not optimization-based.

DiffVG~\cite{diffvg} is a popular library for differentiable 2D rasterization that has been used in many optimization-based SBR methods~\cite{vectorfusion, diffsketcher, svgdreamer}. It supports rendering arbitrary parametric curves, either open or closed, including polygons, ellipses, and polylines. Due to its popularity, we also considered using DiffVG to model Sharpie marker strokes for this work. However, we discovered that out of the box, the DiffVG library does not support rendering polylines that are differentiable with respect to stroke thickness. DiffVG lines are only differentiable with respect to the control points. Furthermore, DiffVG decouples strokes into a boundary shape and a fill color, which we found to be too restrictive because it does not allow us to model the gradual dropoff in darkness from the center of a stroke to the outside. Thus, we choose to implement our own differentiable renderer, Traj2Stroke, which is specialized for rendering polylines.

\vspace{-7pt}\section{Background} \label{sec:background}
Our work is built on FRIDA~\cite{frida}, a robotic system that paints what users describe via text prompts, images, or audio recordings~\cite{robotsynesthesia} in an interactive and collaborative manner~\cite{schaldenbrand2024cofrida}.


\begin{figure}[!tp]
    \centering
    \includegraphics[width=\linewidth]{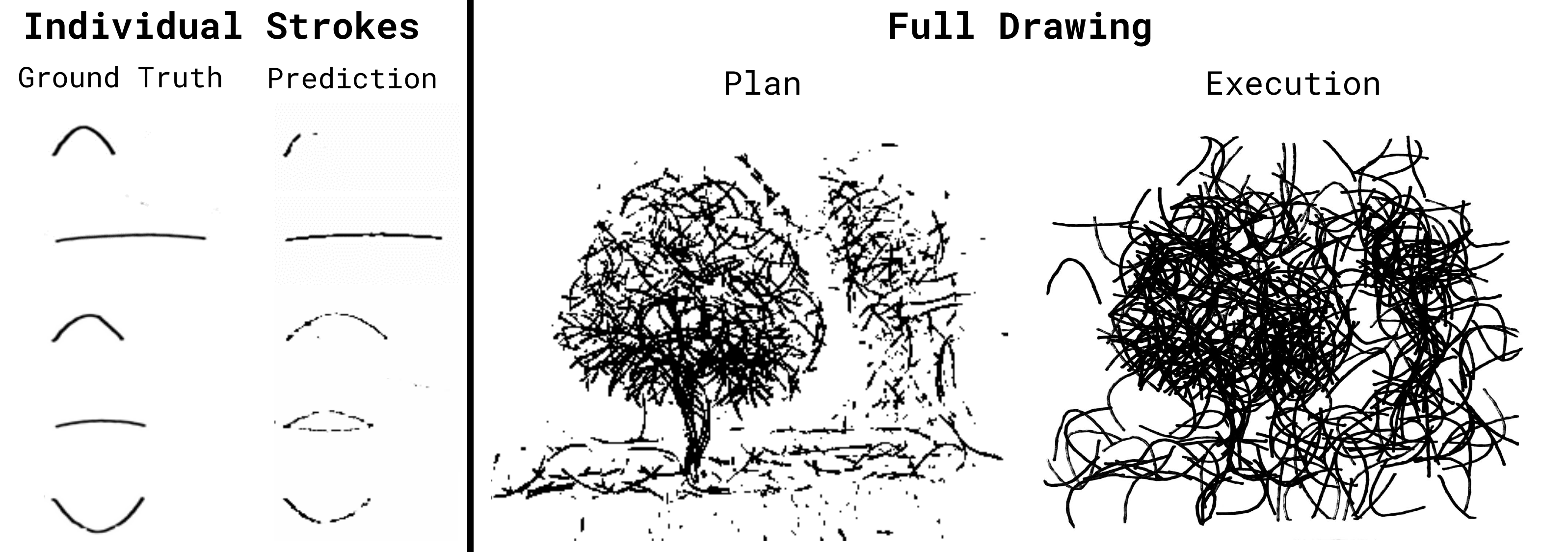}
    \caption{\small \textbf{Weaknesses in FRIDA's stroke model.} Left: FRIDA struggles with rendering thin strokes. Right: for a full drawing, errors in individual stroke predictions accumulate. The optimization process exploits blotchy predictions made by the renderer to create small dots, which appear quite different when actually executed.}
    \label{fig:frida_sim2realgap} \vspace{-15pt}
\end{figure}

FRIDA renders strokes using a convolutional neural network (CNN). Although the CNN works well for short, simple strokes made with a paintbrush, it has trouble converging when longer, thinner Sharpie marker strokes are used, as illustrated in~\Cref{fig:frida_sim2realgap}. We believe this issue arises because thin ground truth strokes require the renderer to achieve a higher level of precision, complicating its task. Consequently, the CNN often produces blotchy, incomplete predictions. This is problematic since the errors compound during the planning stage (illustrated on the right side of~\Cref{fig:frida_sim2realgap}).

Additionally, FRIDA's stroke representation is limited. FRIDA, like many other computer painting works~\cite{schaldenbrand2022styleclipdraw, learningtopaint, stylizedneuralpainting},  uses quadratic Bézier curves defined by three parameters—length, height above the canvas, and bend—to represent strokes. While this straightforward approach offers convenience, it falls short in capturing the complexity and variety inherent in human drawings. In this context, we propose a novel stroke representation and a new renderer to model long, complex trajectories that can enable diverse styles.

\section{Methods}
\subsection{Overview}
Our approach to stroke modeling and rendering consists of (1) capturing and processing human demonstration data using motion capture technology, (2) modeling these trajectories by training an autoencoder, TrajVAE, (3) using Real2Sim2Real methodology to fine-tune our novel rendering approach, Traj2Stroke, and (4) planning using gradient descent to optimize a set of brush stroke parameters through our dynamics model to decrease the feature-space loss between a given image and the predicted painting. 

\begin{figure*}
  \includegraphics[width=\textwidth]{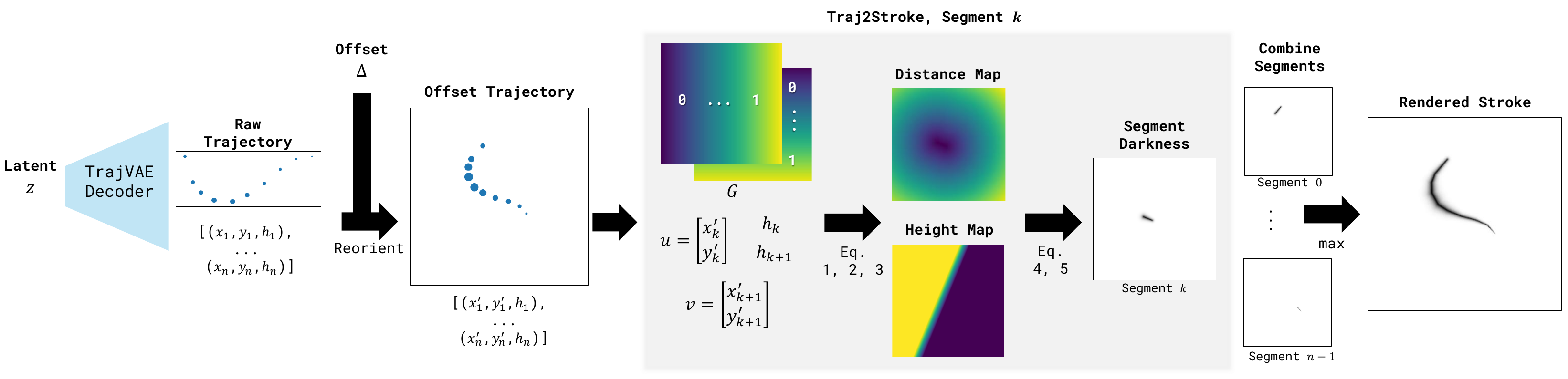}
  \caption{\small \textbf{
  Traj2Stroke.
  } The inputs are a latent vector $z$ and an offset $\Delta$. $z$ is fed through the decoder of a TrajVAE, generating a raw trajectory, which is then rotated and translated according to $\Delta$. We then process the trajectory segments independently, obtaining darkness values for each. Finally, we take the max darkness over all segments.}
  \label{fig:pipeline} \vspace{-7pt}
\end{figure*}



\vspace{-0.7em}
\subsection{Motion Capture Drawing Recording and Processing}
We utilize a motion capture system consisting of OptiTrack cameras and Motive software to capture human brushstroke trajectories.
\begin{figure}[]
    \centering
    \includegraphics[width=\linewidth]{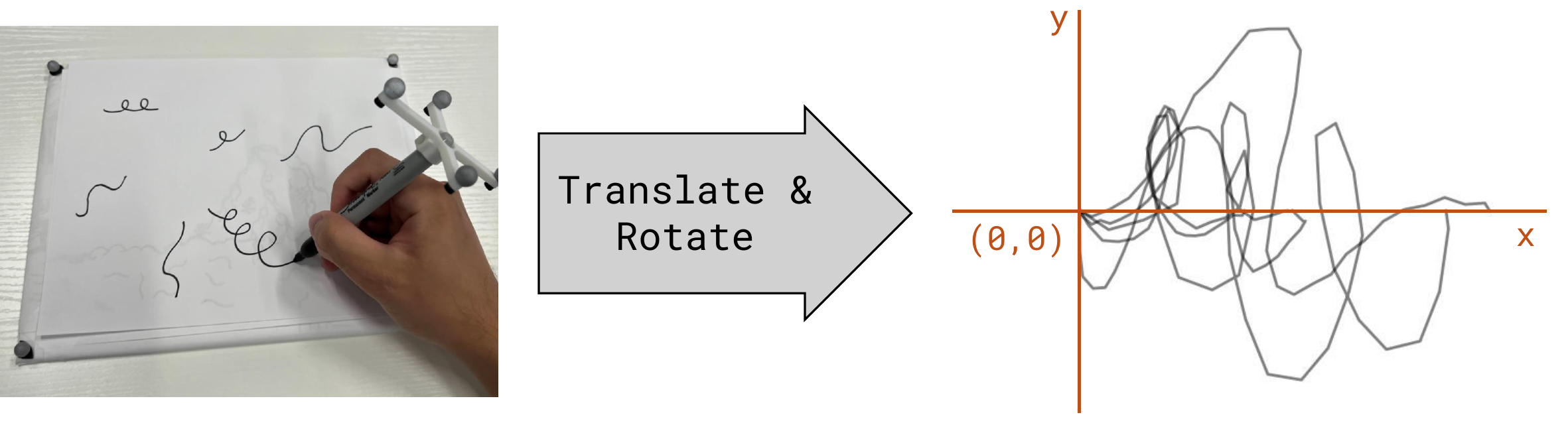}
    \caption{\small \textbf{Mocap setup.} We use a motion capture system to track the position of the canvas and pen over time as an artist draws. Three mocap markers are placed along the corners of the canvas, and four are mounted at the end of the pen. The trajectories of each stroke are extracted, then rotated and translated such that the start of the trajectory is (0, 0) and the end point is on the x-axis.}
    \label{fig:mocap} \vspace{-7pt}
\end{figure}
While the artist sketches, we continuously track the positions and orientations of the canvas and pen. Using a manual measurement of the length of the pen, we are able to calculate the position of the pen tip and determine its distance from the canvas. If this value is below a threshold, we consider the pen to be in contact with the paper. Consecutive positions where this is the case are merged into trajectories. Each trajectory is then standardized by translating it to the origin and rotating it to be horizontal (ending at $y=0$), as seen in \Cref{fig:mocap}, to reduce variation for sample-efficient modeling. It is worth noting that this normalization has a tradeoff: it assumes trajectory style is not affected by position/rotation on the canvas. We also resample each trajectory to have exactly $n$ points (in practice $n=32$).

Thus, each human brushstroke trajectory is modeled as a polyline (piecewise linear) going through $n$ control points. This polyline is encoded as a $n \times 3$ tensor. The coordinates $(x, y, h)$ of each control point are defined by $x$ and $y$ as horizontal displacements (in the plane of the canvas) and $h$ as vertical displacement (elevation of the brush above the canvas).

\vspace{-1em}
\subsection{TrajVAE}
After collecting and processing the motion capture data, we train variational autoencoders \cite{vae} to model these stroke trajectories. We name these TrajVAEs. During training, a TrajVAE takes a trajectory as input, passes it through an MLP encoder that compresses it to a latent vector of size $64$, and then sends it through a MLP decoder to turn it back into a trajectory. We minimize the mean squared error between the input and output trajectories.

We typically record between 20 and 200 human-drawn trajectories per drawing, but found that this is not enough data to robustly train a TrajVAE from scratch. Instead, we pretrain each TrajVAE on trajectories aggregated from multiple recording sessions, then fine-tune it on a single session to capture a more specific style. Each model converges very fast (less than a minute) and only requires a few (\textless20) trajectories in the fine-tuning dataset.

During the planning phase, only the VAE decoder is used. The design of the overall pipeline is modular so that different VAEs can be swapped in, allowing us to change the stroke style with no need for additional training.


Our motion capture device struggles to capture the vertical position of the drawing utensil's tip with enough precision. This is because a small height difference can drastically affect the thickness of a stroke.
Thus, rather than explicitly modeling the height with TrajVAE, in practice we optimize it as separate stroke parameters during the planning process.


\begin{figure*}
    \centering 
    \includegraphics[width=\linewidth]{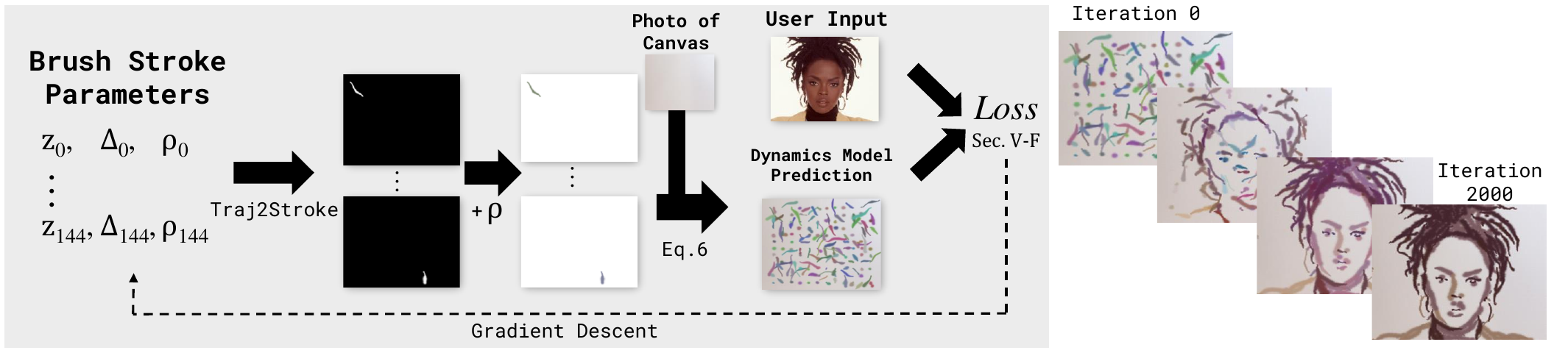}
    \caption{\small 
    \textbf{Planning a Painting.} As described in \mbox{\Cref{sec:planning}}, Spline-FRIDA plans a painting by optimizing the brush stroke parameters through the dynamics model to decrease a features space loss between a given image and the planned painting.
    Whereas FRIDA models brush strokes as simple B\'ezier curves, Spline-FRIDA uses trajectories which enable highly flexible brush strokes.  
    }
    
    \label{fig:planning2}
    \vspace{-1em}
\end{figure*}

\vspace{-0.7em}
\subsection{Traj2Stroke Model \label{traj2stroke_method}}
The TrajVAE model outputs a trajectory that the robot should draw, but to predict the appearance of the stroke given this trajectory, we developed a novel rendering approach that we call Traj2Stroke. Traj2Stroke takes a trajectory, as well as positional and rotational offsets, and renders it as a grayscale image with height and width dimensions, $H\times W$. Importantly, this process is differentiable so that the planning process can backpropagate through it.

To train this model, we randomly sample trajectories from TrajVAE, execute them on the robot, and take before/after pictures of the canvas for each stroke. Next, we input the sampled trajectories into the Traj2Stroke model to get predicted stroke masks. These masks are stamped onto the before-stroke pictures, and the resulting prediction is compared to the after-stroke pictures. We minimize a weighted L1 loss that places higher weight on pixels covered by the new stroke. In practice, collecting the dataset (including setup and execution) takes around an hour, and training takes around 20 minutes to converge on our single-GPU system.

A separate Traj2Stroke model must be trained for each drawing medium (marker/brush), but is robust to out-of-distribution \textit{trajectories}. This means that when we obtain a new TrajVAE, we can almost always plug-and-play it into the system without needing to collect new data and retraining Traj2Stroke. This is very convenient, as collecting the Traj2Stroke dataset is usually the most time-consuming part of preparations.

After receiving a standardized trajectory $[(x_1,y_1,h_1),\cdots,(x_n,y_n,h_n)]$ and pose offsets $\Delta=(\Delta_x,\Delta_y,\Delta_\theta)$, the Traj2Stroke model begins by reorienting the trajectory to be in the reference frame of the canvas (see \Cref{fig:pipeline}). To do this, it first rotates the $x$ and $y$ components by $\Delta_\theta$. Then, each rotated coordinate $(x,y,h)$ is scaled and translated to become
$$\left(m_xx+b_x+\Delta_x,\ m_yy+b_y+\Delta_y,\ h\right).$$
$m_x$, $m_y$, $b_x$, and $b_y$ are learnable parameters used to model any small affine error that may occur during camera calibration. We expect that $m_x,m_y \approx 1$ and $b_x,b_y \approx 0$.

The trajectory has now been converted to canvas coordinates, and we denote it as
$$[(x_1',y_1',h_1),\cdots,(x_n',y_n',h_n)].$$
We proceed by rendering each of its $n-1$ segments separately. Fix an arbitrary $k$, and note that segment $k$ goes from $(x_k',y_k',h_k)$ to $(x_{k+1}',y_{k+1}',h_{k+1})$.

Our approach to rendering the segment is to first define a constant $H\times W\times 2$ tensor $G$ of canvas coordinates, where $H$ and $W$ are the dimensions of the canvas. One channel of this tensor contains the $x$ coordinates, and the other contains the $y$ coordinates, as seen in \Cref{fig:pipeline}. For convenience, we also define $u=\begin{bmatrix}x_k' & y_k'\end{bmatrix}^T$ and $v=\begin{bmatrix}x_{k+1}' & y_{k+1}'\end{bmatrix}^T$.

We compute a \textit{Distance Map} that stores the distance of each coordinate in $G$ to the segment. This is computed with the following equation (note that the vector operations involving $G$ are done element-wise):
\begin{equation}
\begin{split}
    \text{Distance Map} = \min(&\norm{(G-u) - \text{proj}_{v-u}(G-u)},\\
    &\norm{G-u}, \norm{G-v})).
\end{split}
\end{equation}
The first term computes the distance from each point in $G$ to the line through $u$ and $v$, and the last two terms calculate the distance to the endpoints. Thus, taking the minimum of the three yields the distance of each pixel to the line from $u$ to $v$.

We also compute a \textit{Height Map}, which represents the height of the brush tip as it moves over the segment. For each coordinate, we project it onto the segment and compute the height by linear interpolation between $h_k$ and $h_{k+1}$:
\begin{equation}
    T = \text{clamp}_{[0,1]}\left(\frac{\norm{\text{proj}_{v-u}(G-u)}}{\norm{v-u}}\right)
\end{equation}
\begin{equation}
    \text{Height Map} = (1-T)\cdot h_k + T\cdot h_{k+1}.
\end{equation}

We approximate the relationship between the height of the brush tip and the thickness of the stroke as affine. Thus, we introduce two learnable parameters $\alpha$ and $\beta$, and obtain a \textit{Thickness Map} like so:
\begin{equation}
    \text{Thickness Map} = \alpha\cdot\text{Height Map} + \beta.
\end{equation}

If the distance between a coordinate and the segment is less than the stroke thickness, then that coordinate should be affected by the stroke. We assume there is a gradual dropoff in darkness as we get further from the center of the segment. This reasoning motivates the following calculation for the darkness values:
\begin{equation}
    \text{Darkness} = \left[\text{clamp}_{[0,1]}\left(1-\frac{\text{Distance Map}}{\text{Thickness Map}}\right)\right]^c.
\end{equation}
Coordinates directly on the segment get a darkness value of $1$, and coordinates that are a stroke thickness away get a darkness value of $0$. This also introduces another learnable parameter $c$ which determines how quickly the darkness values drop off as they get further from the segment.

Finally, we take the max darkness values over all segments to obtain the rendered stroke.

In total, the Traj2Stroke model has only $7$ learnable parameters: $m_x$, $m_y$, $b_x$, $b_y$, $\alpha$, $\beta$, and $c$.

\subsection{Stroke Composition} \label{sec:dyanmics_model}

We define each brush stroke action as a set of parameters: TrajVAE latent vector $z$, pose offsets $\Delta$, and RGB color $\rho$. 
Given $z$ and $\Delta$,~\mbox{\Cref{fig:pipeline}} illustrates how Spline-FRIDA predicts the shape of a single stroke. Next, the rendered stroke is colorized by duplicating it to 3-channels and multiplying each channel based on the stroke color as seen in~\mbox{\Cref{fig:planning2}}.
We can predict how this stroke $s$ will appear once it is performed on a canvas $c_t$ by stamping it via an alpha blending formula.


\subsection{{Painting and Drawing Planning}} \label{sec:planning}

To plan a painting or drawing, we follow the FRIDA~\mbox{\cite{frida}} planning algorithm which plans paintings using an optimization loop, depicted in~\mbox{\Cref{fig:planning2}}. A user-specified number of brush stroke actions are randomly initialized. 
At each optimization step, the current canvas is compared to the user-specified target image forming a loss value. In practice, features from the planned painting and target image are extracted using pretrained neural networks (e.g., CLIP~\mbox{\cite{radford2021-clip}}) and compared using cosine similarity as introduced in~\mbox{\cite{vinker2022clipasso}}. The loss is back-propagated through the dynamics model to the brush stroke parameters which are updated using gradient descent.

If the robot is painting in color, the color parameters are optimized as continuous RGB values during initial iterations.
In the last 10\% of optimization iterations, the algorithm discretizes the colors to a user-specified number using K-Means clustering. After optimizing for 2000 iterations, the system shows which colors of paint need to be mixed in a graphical user-interface. The user mixes these paint colors, provides them to the robot, then the robot can begin painting.




\section{Results\label{results}}
\begin{figure*}[!h]
    \centering
    \includegraphics[width=0.9\textwidth]{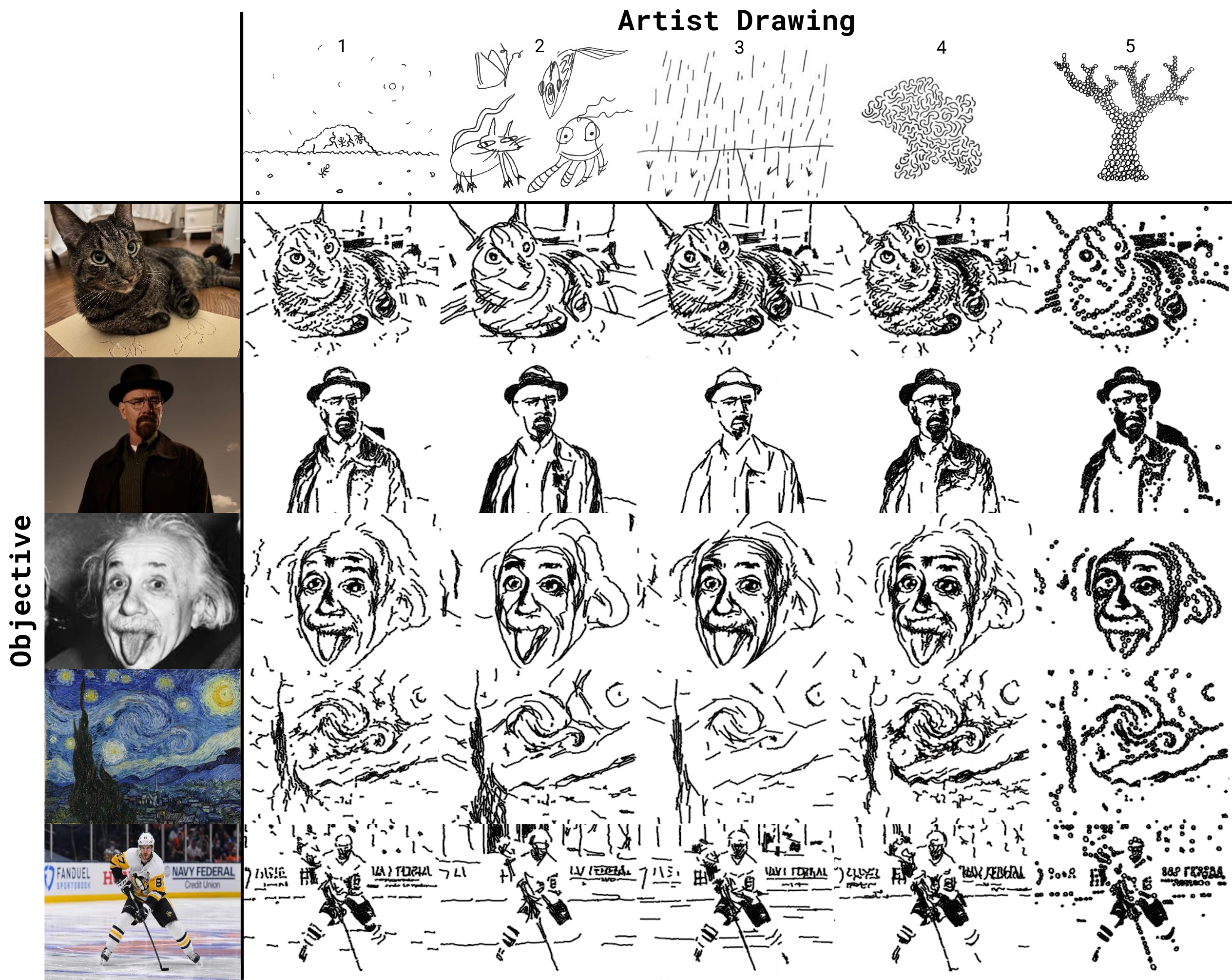}
    \caption{\small \textbf{Example drawings made by Spline-FRIDA.} Each column represents a distinct trajectory style and each row uses a different objective. 
    The top row contains original drawings made by human artists on our mocap system. One VAE was fine-tuned on each human drawing and used to plan the drawings in each column.}
    \label{fig:full_pipeline}
    \vspace{-1em}
\end{figure*}

\Cref{fig:full_pipeline} shows an array of drawings produced by Spline-FRIDA. We hand-pick five human drawings from members of our lab using the mocap system, each with distinct stroke styles, which are presented in the top row. Each human drawing is used to fine-tune a separate TrajVAE, resulting in five unique TrajVAEs. Each TrajVAE is then used to plan a series of drawings with various objectives. These objectives are displayed in the left column.

The individual styles of the drawing trajectories are preserved by the TrajVAEs. For instance, the fourth human drawing exhibits tiny, curly lines, which are reproduced in the drawings made using its corresponding TrajVAE. Similarly, the fifth human drawing is composed of small circles, which is also true for the robot drawings in its column.

\vspace{-1em}
\subsection{Human Evaluations}
To what extent is Spline-FRIDA able to capture the stroke style of a drawing? And, in general, are Spline-FRIDA's drawings better than those made by FRIDA? These questions are subjective and difficult to answer with automatic metrics. To obtain quantitative results, we conducted a survey and released it to 100 participants on Amazon Mechanical Turk.

\begin{figure}[]
    \centering
    \vspace{-1em}
    \includegraphics[width=0.8\columnwidth]{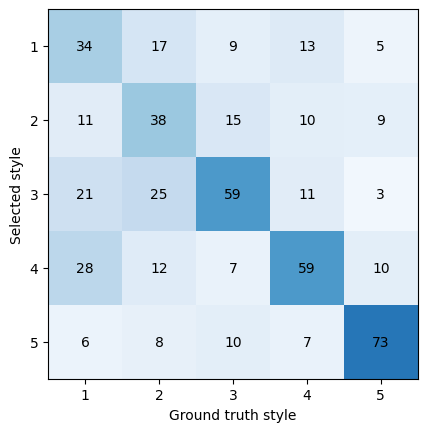}
    \caption{\small \textbf{Confusion matrix for matching task.} The x-axis represents the index of the specific TrajVAE used to generate the drawing, and the y-axis represents the index of the human drawing participants thought was most similar. The five human drawings/styles the same ones as in the top row of \Cref{fig:full_pipeline}, with the same order.}
    \label{fig:matching}
\end{figure}

For the first part of the survey, we asked participants to match Spline-FRIDA drawings with human drawings that have the same stroke style. More specifically, for each participant, we selected a random human drawing, along with five robot drawings (a random row of \Cref{fig:full_pipeline}), and asked them to pick the robot drawing that best matched the style of the human drawing. We told participants to ``focus on the characteristics of individual strokes, such as their trajectories, shapes, and curves.'' The results of this experiment are seen in \Cref{fig:matching}.

The high values of \Cref{fig:matching} along the diagonal suggest that, in general, participants were able to choose the correct human drawing used to style each robot drawing. Style 5 seemed to be particularly distinguishable. Meanwhile, style 1 was often confused with style 4, and style 2 was confused with style 3. Nevertheless, all five encoded styles are most strongly associated with the correct human drawings.

The second part of the survey asked participants' subjective opinions on Sharpie drawings made by FRIDA vs. Spline-FRIDA. Each participant was shown an objective image and two robot drawings of it, one from FRIDA and one from Spline-FRIDA. Both robot drawings were executed on the physical robot so that any Sim2Real gap comes into play. The questions and tallied responses are shown in \Cref{tab:opinions}.

\begin{table}[]
    \centering
    \begin{tabular}{p{0.5\linewidth}lcc@{}}
    \toprule
     & FRIDA & Spline-FRIDA \\ \midrule
    Which drawing looks more like it was drawn by a human (rather than a robot)? & 27 & 73 \\ \midrule
    Which drawing looks better overall? & 16 & 84 \\ \midrule
    Which drawing better matches the reference image? & 16 & 84 \\ \midrule
    Which drawing is more artistic? & 18 & 82 \\ \midrule
    Which drawing is more abstract? & 40 & 60 \\ \bottomrule
    \end{tabular}
    \caption{\small \textbf{Opinions on FRIDA vs Spline-FRIDA.} Each cell shows the number of participants that chose the system for the given question. Overall, participants thought that compared to the B\`ezier curve representation of FRIDA, drawings made by Spline-FRIDA were more human-like, higher quality, more true to the objective, and more artistic.}
    \label{tab:opinions} \vspace{-1em}
\end{table}

\begin{figure}
    \centering
    \includegraphics[width=0.9\linewidth]{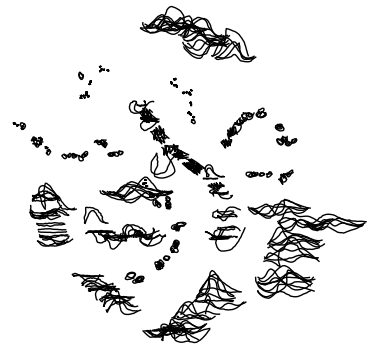}
    \caption{\small \textbf{Mapping the latent space.} We visualize the TrajVAE latent space by drawing trajectories at their respective coordinates, projected down to 2 dimensions via t-SNE. To generate this plot, we use a TrajVAE that is trained on multiple sessions of human trajectory data.}
    \label{fig:tsne}
    \vspace{-1em}
\end{figure}

Respondents believed that Spline-FRIDA's drawings, in comparison to FRIDA's, appeared more human-made, had higher overall quality, better matched the reference image, and were more artistic. Respondents also perceived the Spline-FRIDA drawings as more ``abstract'', although opinions on this were somewhat split.

\vspace{-1em}
\subsection{Trajectory Distributions}
In 
\mbox{\Cref{fig:tsne}}, we visualize the latent space for our TrajVAE trained on multiple drawing sessions. We encode all of the human trajectories into latent vectors, then project them down to 2 dimensions using t-SNE \cite{tsne}. We then draw each human trajectory at its corresponding 2d coordinates.
We observe human trajectories spread throughout the space, forming several homogeneous clusters. This structured organization indicates that TrajVAE effectively learns a correlation between trajectories and latent vectors. Consequently, an optimization-based planning algorithm is likely to be effective.

\vspace{-0.8em}
\subsection{Brush Stroke Dynamics Modeling Experiments\label{traj2stroke_improvements}}
The purpose of a stroke dynamics model is to differentiably render trajectories. We experiment with a variety of methods to do this and evaluate them both quantitatively and qualitatively in a controlled experiment. In total, we evaluated four methods:

\subsubsection{CNN} Our baseline, a convolutional neural network starting with a fully connected layer and followed by several transposed convolutions. This is analogous to FRIDA's~\cite{frida} renderer architecture, except it takes in full trajectory rather than the parameterization (length, bend, height) that FRIDA uses.
\subsubsection{CNN with CoordConv} To render a trajectory, one subproblem the renderer must solve is mapping Cartesian coordinates to one-hot pixel space. Liu et. al. \cite{coordconv} showed that traditional CNNs can have difficulty with this, so we implement their suggestion of using CoordConv layers instead of traditional convolutions. This means adding two additional channels to the input of each convolution: one containing the x-coordinates of each pixel, and the other containing the y-coordinates.
\subsubsection{Traj2Stroke} This is our main method, with the rule-based transformations, that was described in Section \ref{traj2stroke_method}.
\subsubsection{Traj2Stroke with U-Net} Our main method, but with an additional convolutional network attached after the output layer. The goal of this additional network is to refine the Traj2Stroke output by learning subtle effects such as texture and bristle drag. Its architecture closely follows that of U-Net \cite{unet}. We freeze the U-Net weights during the first half of the training and unfreeze them for the second half. The purpose of this is to train the Traj2Stroke portion first and get it as close as possible to the ground truth, before using the U-Net to refine it. Inspired by the success of ControlNet \cite{controlnet}, the U-net is initialized with a zero-convolution final transformation so that it initially performs the identity function.

Since generating training strokes and training a new dynamics model for every new stroke style is time consuming, the ability of the dynamics model to generalize to unseen styles is important.
%
%
In order to evaluate generalizability, we train and test the stroke model on trajectories from different distributions. More precisely, we create two datasets, A and B. Both datasets contain (trajectory, stroke image) pairs. For dataset A, the trajectories are sampled from a generic TrajVAE, trained on a session that we judge to have good stroke diversity. For dataset B, we use trajectories from more specialized TrajVAEs, trained on sessions with very unique styles. We train the model using dataset A, and we evaluate generalizability by checking its performance on dataset B.

We run the experiment twice, once for each of two drawing mediums: a sharpie and a thin paintbrush. The experiment results can be seen in \Cref{tab:stroke_model_comparisons}. The Traj2Stroke architecture without U-Net achieves the lowest loss on sharpie strokes. Adding the U-Net hurts performance on sharpie strokes, though it achieves the best results on brush strokes. There is not a substantial increase in performance from using CoordConv over traditional the pure CNN architecture.

\begin{table}[!tp]
    \centering
    \begin{tabular}{@{}lcccc@{}}
    \toprule
    Medium & CNN & \begin{tabular}[c]{@{}c@{}}CNN\\ w/ CoordConv\end{tabular} & Traj2Stroke & \begin{tabular}[c]{@{}c@{}}Traj2Stroke\\ w/ U-Net\end{tabular} \\ \midrule
    Sharpie & .00107 & .00095 & \textbf{.00055} & .00098 \\
    Brush & .00162 & .00163 & .00158 & \textbf{.00153} \\ \bottomrule
    \end{tabular}
    \vspace{1em}
    \caption{\small \textbf{Quantitative comparison of stroke models.} This table shows the average L1 loss of each stroke model when predicting either sharpie or brush strokes (lower is better). Loss is calculated on dataset B (out-of-distribution) trajectories only. Traj2Stroke achieves the best results for sharpie strokes, and Traj2Stroke with U-Net is the best for brush strokes.}
    \label{tab:stroke_model_comparisons}
\end{table}

\begin{figure}[!tp]
    \centering
    \vspace{-1em}
    \includegraphics[width=\linewidth]{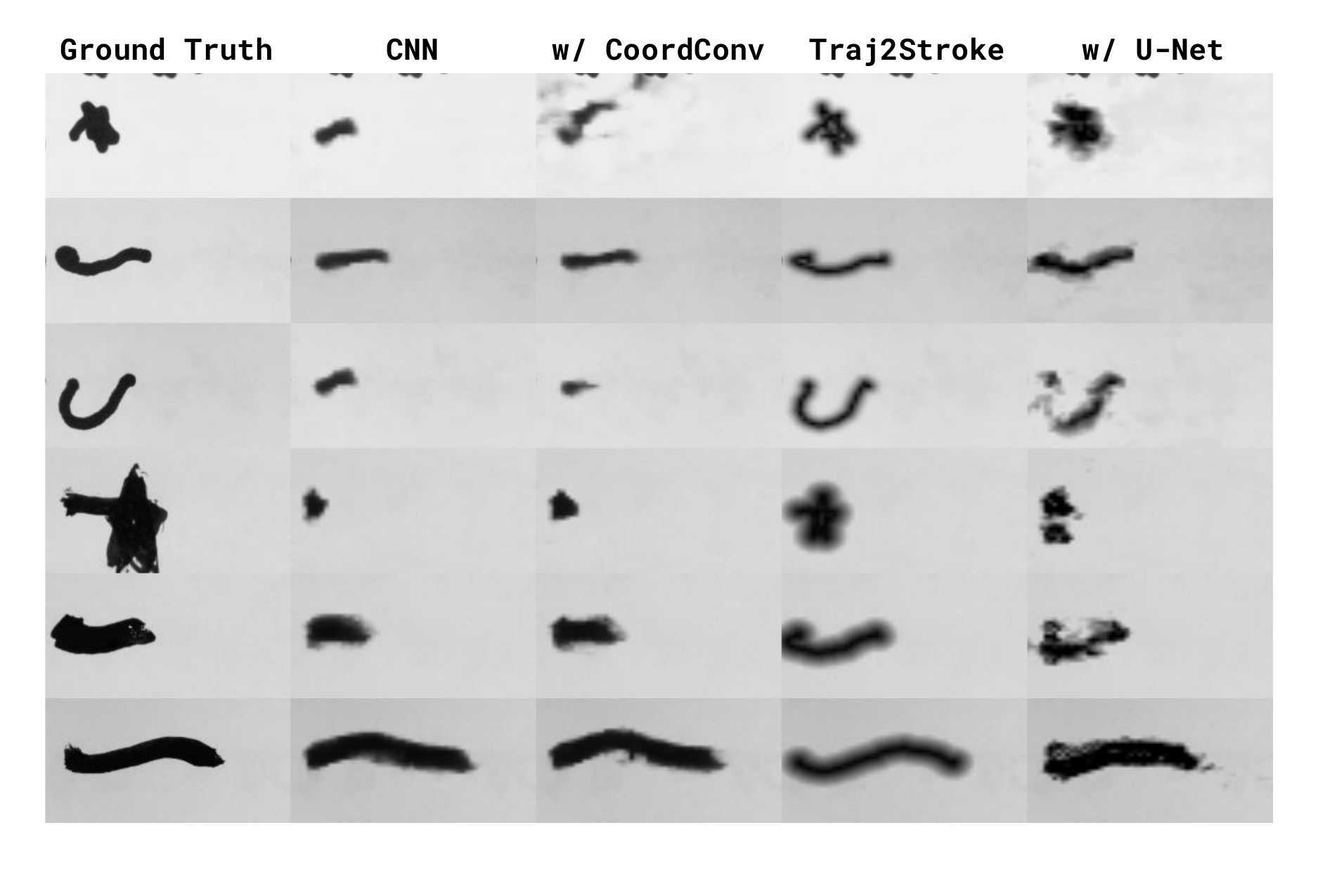}
    \vspace{-2.5em}
    \caption{\small \textbf{Visualizing the outputs of various stroke models.} The first three rows contain sharpie strokes, and the last three contain brush strokes all of which were made from samples using TrajVAE models not used for training. 
    }
    \label{fig:stroke_model_comparisons}
    \vspace{-1em}
\end{figure}

Visually, example predictions generated by each model can be seen in \Cref{fig:stroke_model_comparisons}. All examples are from dataset B, meaning that these trajectories 
are out of distribution from the training set.
The vanilla CNN with and without CoordConv fails to generalize in certain cases. 
The Traj2Stroke model performs near-perfect for the sharpie strokes and captures the general shape of the brush strokes. Adding the U-Net to Traj2Stroke helped capture the texture of the brush strokes, but the added parameters hurt generalization to very out-of-distribution strokes, such as the star example. 

\begin{figure}[!htp]
    \centering
    \includegraphics[width=\linewidth]{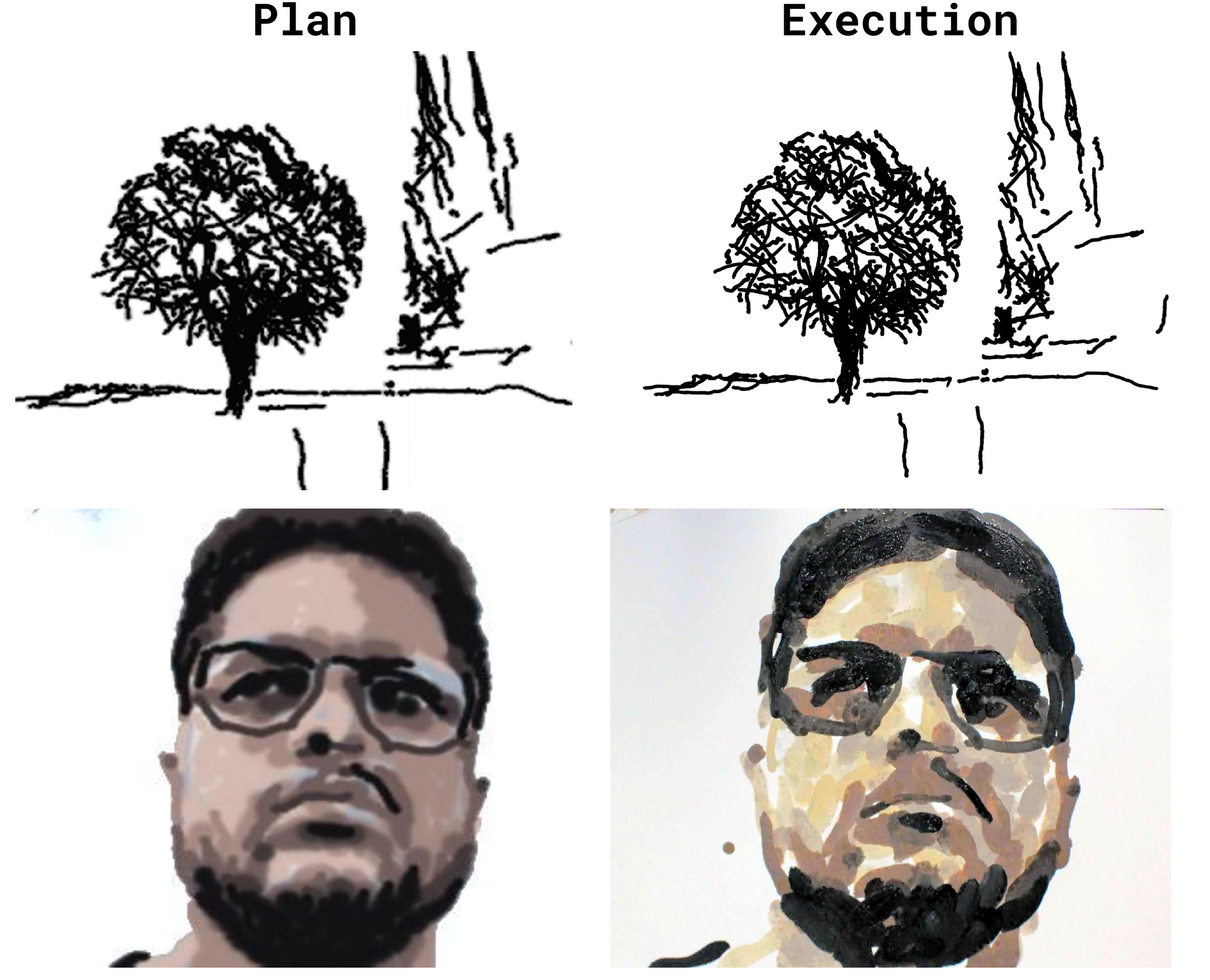}
    \caption{\small \textbf{Spline-FRIDA's low Sim2Real gap.} We compare a plan made by Spline-FRIDA with its execution (physically drawn with a robot). The top row with a black marker, and the bottom row with a paint brush.}
    \label{fig:splinefrida_sim2realgap}
    \vspace{-1em}
\end{figure}

Based on these findings, we choose to implement the base Traj2Stroke model (without U-Net) for Spline-FRIDA. As illustrated in~\Cref{fig:splinefrida_sim2realgap}, the resulting Sim2Real gap for Sharpie drawings is very low. This is a huge improvement compared to the original FRIDA results depicted in~\Cref{fig:frida_sim2realgap}.

\section{Limitations/Future Work}

In this paper, we focus on capturing style by modeling brush stroke trajectories. However, these are intra-stroke attributes and we do not consider other aspects of style, including inter-stroke style elements.

Additionally, Spline-FRIDA is computationally expensive (400-stroke paintings take approximately one hour to plan for on a machine with an NVIDIA GeForce 4090). Hardware memory constraints also force us to optimize only a subset (80 in practice) of strokes at a time. Beyond computational inneficiencies, the optimization methods for  stroke-based rendering in general are susceptible to local minima. Results are dependent on stroke parameter initialization. In future work, we would like to explore training models, such as reinforcement learning agents, to perform stroke prediction. This could improve runtime planning speeds, memory constraints, and potentially help avoid local minima.

\section{Conclusions}

In this paper, we introduce Spline-FRIDA, a method to enable a human user to create sketches and paintings using a robot in a diverse range of stroke styles. Using motion capture, we collect datasets of human stroke trajectories in multiple styles, which we capture with variational autoencoders. We find that the existing stroke dynamics model fails to render these complex strokes, so we introduce a novel method, Traj2Stroke. Our evaluations with human subjects indicate that Spline-FRIDA effectively captures the style of human-created illustrations and produces renderings that more closely align with the reference image than the previous model. Our findings also reveal that these flexible strokes lend a more artistic and human-like appearance, distinguishing them from typical robotic creations.
\section{Acknowledgements}
We thank Howard Halim for his early ideas related to the Traj2Stroke model. This work was supported in part by the Technology Innovation Program (20018295, Meta-human: a virtual cooperation platform for specialized industrial services) and funded by the Ministry of Trade, Industry \& Energy (MOTIE, Korea) and NSF IIS-2112633.

\footnotesize
\bibliographystyle{IEEEbib}
\bibliography{refs}

\begin{thebibliography}{10}

\bibitem{robertson1967-formAndContent}
Duncan Robertson,
\newblock ``The dichotomy of form and content,''
\newblock {\em College English}, vol. 28, no. 4, pp. 273--279, 1967.

\bibitem{schaldenbrand2022styleclipdraw}
Peter Schaldenbrand, Zhixuan Liu, and Jean Oh,
\newblock ``Styleclipdraw: Coupling content and style in text-to-drawing translation,''
\newblock {\em arXiv preprint arXiv:2202.12362}, 2022.

\bibitem{gebru2023artAiImpact}
Harry~H. Jiang, Lauren Brown, Jessica Cheng, Mehtab Khan, Abhishek Gupta, Deja Workman, Alex Hanna, Johnathan Flowers, and Timnit Gebru,
\newblock ``Ai art and its impact on artists,''
\newblock in {\em Proceedings of the 2023 AAAI/ACM Conference on AI, Ethics, and Society}, New York, NY, USA, 2023, AIES '23, p. 363–374, Association for Computing Machinery.

\bibitem{jansen2021coDrawWorkflows}
Chipp Jansen and Elizabeth Sklar,
\newblock ``Exploring co-creative drawing workflows,''
\newblock {\em Frontiers in Robotics and AI}, vol. 8, pp. 577770, 2021.

\bibitem{bateman2021creating}
Cole Bateman,
\newblock {\em Creating for Creatives: A Humanistic Approach to Designing AI Tools Targeted at Professional Animators},
\newblock Ph.D. thesis, Harvard University, 2021.

\bibitem{davis2016DrawingApprentice}
Nicholas Davis, Chih-PIn Hsiao, Kunwar Yashraj~Singh, Lisa Li, and Brian Magerko,
\newblock ``Empirically studying participatory sense-making in abstract drawing with a co-creative cognitive agent,''
\newblock in {\em Proceedings of the 21st International Conference on Intelligent User Interfaces}, 2016, pp. 196--207.

\bibitem{lawton2023reframer}
Tomas Lawton, Francisco~J Ibarrola, Dan Ventura, and Kazjon Grace,
\newblock ``Drawing with reframer: Emergence and control in co-creative ai,''
\newblock in {\em Proceedings of the 28th International Conference on Intelligent User Interfaces}, 2023, pp. 264--277.

\bibitem{lee2024adversarialRobot}
Shayla Lee and Wendy Ju,
\newblock ``Adversarial robots as creative collaborators,''
\newblock {\em arXiv preprint arXiv:2402.03691}, 2024.

\bibitem{learningtopaint}
Zhewei Huang, Wen Heng, and Shuchang Zhou,
\newblock ``Learning to paint with model-based deep reinforcement learning,'' 2019.

\bibitem{frida}
Peter Schaldenbrand, James McCann, and Jean Oh,
\newblock ``Frida: A collaborative robot painter with a differentiable, real2sim2real planning environment,'' 2022.

\bibitem{stylizedneuralpainting}
Zhengxia Zou, Tianyang Shi, Shuang Qiu, Yi~Yuan, and Zhenwei Shi,
\newblock ``Stylized neural painting,'' 2020.

\bibitem{contentmaskedloss}
Peter Schaldenbrand and Jean Oh,
\newblock ``Content masked loss: Human-like brush stroke planning in a reinforcement learning painting agent,'' 2021.

\bibitem{painttransformer}
Songhua Liu, Tianwei Lin, Dongliang He, Fu~Li, Ruifeng Deng, Xin Li, Errui Ding, and Hao Wang,
\newblock ``Paint transformer: Feed forward neural painting with stroke prediction,'' 2021.

\bibitem{layerwise}
Xu~Ma, Yuqian Zhou, Xingqian Xu, Bin Sun, Valerii Filev, Nikita Orlov, Yun Fu, and Humphrey Shi,
\newblock ``Towards layer-wise image vectorization,'' 2022.

\bibitem{helpinghand}
Jingwan Lu, Fisher Yu, Adam Finkelstein, and Stephen DiVerdi,
\newblock ``Helpinghand: example-based stroke stylization,''
\newblock {\em ACM Trans. Graph.}, vol. 31, no. 4, jul 2012.

\bibitem{handwriting}
Tom S.~F. Haines, Oisin Mac~Aodha, and Gabriel~J. Brostow,
\newblock ``My text in your handwriting,''
\newblock {\em ACM Trans. Graph.}, vol. 35, no. 3, may 2016.

\bibitem{diffvg}
Tzu-Mao Li, Michal Luk\'{a}\v{c}, Gharbi Micha\"{e}l, and Jonathan Ragan-Kelley,
\newblock ``Differentiable vector graphics rasterization for editing and learning,''
\newblock {\em ACM Trans. Graph. (Proc. SIGGRAPH Asia)}, vol. 39, no. 6, pp. 193:1--193:15, 2020.

\bibitem{vectorfusion}
Ajay Jain, Amber Xie, and Pieter Abbeel,
\newblock ``Vectorfusion: Text-to-svg by abstracting pixel-based diffusion models,'' 2022.

\bibitem{diffsketcher}
XiMing Xing, Chuang Wang, Haitao Zhou, Jing Zhang, Qian Yu, and Dong Xu,
\newblock ``Diffsketcher: Text guided vector sketch synthesis through latent diffusion models,''
\newblock in {\em Advances in Neural Information Processing Systems}, A.~Oh, T.~Naumann, A.~Globerson, K.~Saenko, M.~Hardt, and S.~Levine, Eds. 2023, vol.~36, pp. 15869--15889, Curran Associates, Inc.

\bibitem{svgdreamer}
Ximing Xing, Haitao Zhou, Chuang Wang, Jing Zhang, Dong Xu, and Qian Yu,
\newblock ``Svgdreamer: Text guided svg generation with diffusion model,''
\newblock in {\em Proceedings of the IEEE/CVF Conference on Computer Vision and Pattern Recognition (CVPR)}, June 2024, pp. 4546--4555.

\bibitem{robotsynesthesia}
Vihaan Misra, Peter Schaldenbrand, and Jean Oh,
\newblock ``Robot synesthesia: A sound and emotion guided ai painter,'' 2023.

\bibitem{schaldenbrand2024cofrida}
Peter Schaldenbrand, Gaurav Parmar, Jun-Yan Zhu, James McCann, and Jean Oh,
\newblock ``Cofrida: Self-supervised fine-tuning for human-robot co-painting,''
\newblock in {\em 2024 IEEE International Conference on Robotics and Automation (ICRA)}. IEEE, 2024.

\bibitem{vae}
Diederik~P Kingma and Max Welling,
\newblock ``Auto-encoding variational bayes,'' 2022.

\bibitem{radford2021-clip}
Alec Radford, Jong~Wook Kim, Chris Hallacy, Aditya Ramesh, Gabriel Goh, Sandhini Agarwal, Girish Sastry, Amanda Askell, Pamela Mishkin, Jack Clark, Gretchen Krueger, and Ilya Sutskever,
\newblock ``Learning transferable visual models from natural language supervision,''
\newblock {\em CoRR}, vol. abs/2103.00020, 2021.

\bibitem{vinker2022clipasso}
Yael Vinker, Ehsan Pajouheshgar, Jessica~Y Bo, Roman~Christian Bachmann, Amit~Haim Bermano, Daniel Cohen-Or, Amir Zamir, and Ariel Shamir,
\newblock ``Clipasso: Semantically-aware object sketching,''
\newblock {\em arXiv preprint arXiv:2202.05822}, 2022.

\bibitem{tsne}
Laurens van~der Maaten and Geoffrey Hinton,
\newblock ``Visualizing data using {t-SNE},''
\newblock {\em Journal of Machine Learning Research}, vol. 9, pp. 2579--2605, 2008.

\bibitem{coordconv}
Rosanne Liu, Joel Lehman, Piero Molino, Felipe~Petroski Such, Eric Frank, Alex Sergeev, and Jason Yosinski,
\newblock ``An intriguing failing of convolutional neural networks and the coordconv solution,'' 2018.

\bibitem{unet}
Olaf Ronneberger, Philipp Fischer, and Thomas Brox,
\newblock ``U-net: Convolutional networks for biomedical image segmentation,'' 2015.

\bibitem{controlnet}
Lvmin Zhang, Anyi Rao, and Maneesh Agrawala,
\newblock ``Adding conditional control to text-to-image diffusion models,'' 2023.

\end{thebibliography}
\end{document}